\UseRawInputEncoding

\documentclass[letterpaper]{article} 
\usepackage[]{aaai24}  
\usepackage{times}  
\usepackage{helvet}  
\usepackage{courier}  
\usepackage[hyphens]{url}  
\usepackage{graphicx} 
\def\UrlFont{\rm}  
\usepackage{natbib}  
\usepackage{caption} 
\usepackage{algorithm}
\usepackage{algorithmic}

\usepackage{tabularx}
\usepackage{booktabs}
\usepackage{multirow}
\usepackage{amsmath}
\usepackage{amssymb}
\usepackage{xcolor}

\usepackage{newfloat}
\usepackage{listings}
\DeclareCaptionStyle{ruled}{labelfont=normalfont,labelsep=colon,strut=off} 
\floatstyle{ruled}
\newfloat{listing}{tb}{lst}{}
\floatname{listing}{Listing}
\title{MF-NeRF: Memory Efficient NeRF with Mixed-Feature Hash Table}
\author {
    Yongjae Lee\textsuperscript{\rm 1},
    Li Yang\textsuperscript{\rm 2},
    Deliang Fan\textsuperscript{\rm 1}
}
\affiliations {
    \textsuperscript{\rm 1}Johns Hopkins University\\
    \textsuperscript{\rm 2}University of North Carolina at Charlotte\\
    ylee236@jhu.edu, lyang50@charlotte.edu, dfan10@jhu.edu
}

\usepackage{bibentry}

\begin{document}

\maketitle

\begin{abstract}
    Neural radiance field (NeRF) has shown remarkable performance in generating photo-realistic novel views. 
    Among recent NeRF related research, the approaches that involve the utilization of explicit structures like grids to manage features achieve exceptionally fast training by reducing the complexity of multilayer perceptron (MLP) networks.
    However, 
    storing features in dense grids demands a substantial amount of memory space, resulting in a notable memory bottleneck within computer system. Consequently, it leads to a significant increase in training times without prior hyper-parameter tuning.
    To address this issue, in this work, we are the first to propose MF-NeRF, a memory-efficient NeRF framework that employs a Mixed-Feature hash table to improve memory efficiency and reduce training time while maintaining reconstruction quality. Specifically, we first design a \textit{mixed-feature hash encoding} to adaptively mix part of multi-level feature grids and map it to a single hash table. Following that, in order to obtain the correct index of a grid point, we further develop an \textit{index transformation} method that transforms indices of an arbitrary level grid to those of a canonical grid. Extensive experiments benchmarking with state-of-the-art Instant-NGP, TensoRF, and DVGO, indicate our MF-NeRF could achieve the fastest training time on the same GPU hardware with similar or even higher reconstruction quality. 
\end{abstract}

\section{Introduction}
Representing 3D scenes has garnered significant recognition in various industries. In the past, the conventional method involved breaking down free surfaces into primitive triangles, also known as polygons. However, this approach has been proven to be costly, so it often simplifies the scene  
to reduce the cost which in turn leads to a sacrifice in the quality of reconstruction.
Recently, the developments of machine learning have brought new possibilities for scene representation \cite{Niemeyer2020Differentiable,Park2019DeepSDF}. Notably, the neural radiance field (NeRF) \cite{Mildenhall2020NeRF} has made a significant advancement in both the quality and efficiency of synthesizing novel views, which learns a 3D scene by training multi-layer perceptrons (MLPs).  

\begin{figure}[!t]
    \centering
    \includegraphics[width=\columnwidth]{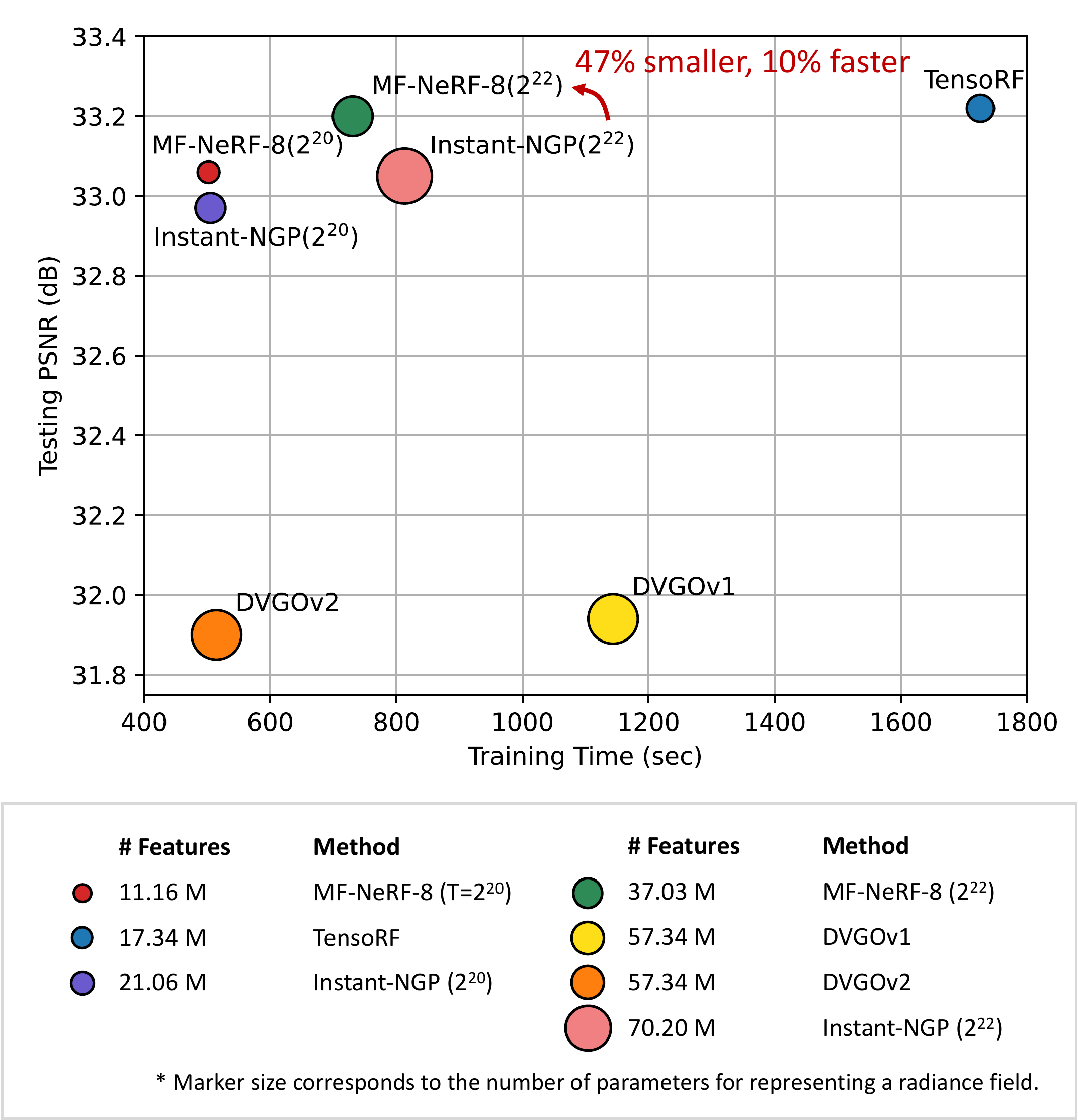}
    \caption{Comparison of training time and rendering quality on the Synthetic NeRF dataset. Our MF-NeRF efficiently represents scenes using less parameters, enhanced by sharing features across multi-resolution grids. Remarkably, it uses only half of the parameters, and less training time, as Instant-NGP with the same hash table size $T$. 
    }
    \label{fig:1}
\end{figure}

In general, NeRF has two main research directions in encoding a 3D scene: implicit and explicit representations. In the implicit representation-based methods \cite{Guo2022Incremental,Martin2021NeRFintheWild,Schwarz2020GRAF,Hong2022HeadNeRF,Turki2022MegaNeRF,Chen2022AugNeRF}, it is crucial to provide inputs of sufficient size to encode high-dimensional and complex radiance fields into MLP networks. As one common approach, the positional encoding \cite{Mildenhall2020NeRF,Vaswani2017Attention} method decomposes low-dimensional spatial-directional inputs
into a set of high-dimensional frequency parameters, helping MLP networks 
learn complex curves and view-dependent light effects. 
For processing high-dimensional inputs effectively, the structure of MLP networks needs to be both deep and wide.
Additionally, due to the infinite and continuous nature of the space, a multitude of giant MLP network queries are required to learn every direction through every point. As a result, the training time and rendering time are typically very large.

Differently, other prior works \cite{Liu2020Neural,Yu2021PlenOctrees,Takikawa2021Neural,Chen2022GeometryGuided,Hedman2021Baking} address the issue of MLP complexity by abstracting and managing scenes as features through explicit representation. They mainly utilize a feature grid that can inherently represent spatial information through voxel coordinates, enabling the design of small and efficient MLPs. While these approaches drastically reduce the computing time for learning and rendering, they typically require enormous GPU memory for storing the large, dense feature grids. To address such memory bottleneck challenge, 
feature compression methods have been discussed\cite{Chen2022TensoRF,Chen2022TensoRF}. As a representative work, Instant-NGP \cite{mueller2022instant} proposes
multi-resolution hash table to manage multiple grids of different resolutions and then compress them by mapping into smaller hash tables.
However, having more features in the grid requires a larger allocation of memory space for the big hash tables and cache buffering in memory. If such memory space is much larger than computer cache memory, the training time will increase significantly due to greatly increased cache miss ratio, without a corresponding improvement in rendering quality, as shown in the analysis of Instant-NGP with hash table sizes ranging from $2^{20}$ to $2^{22}$ in \ref{fig:1}. In real applications, this memory bottleneck issue can arise as a major problem when learning on large scenes, such as buildings \cite{Tancik2022BlockNeRF}.

To address such challenge, following the explicit representation approach, we propose MF-NeRF, a memory efficient NeRF framework with a novel \textbf{Mixed-Feature (MF) hash encoding} method to reduce memory usage and training time, while preserving the quality of reconstruction. Our proposed MF hash encoding maps multiple feature grids into a shared MF hash table, enabling the learning of combined/mixed coarse and fine abstracted features among grids. 
As a consequence, MF-NeRF could significantly compress the conventional multiresolution hash feature encoding, leading to a great reduction in training time while maintaining high rendering quality (e.g., MF-NeRF-8 ($2^{22}$) outperforms Instant-NGP ($2^{22}$) on the same GPU with 47\% smaller model size and 10\% reduction in training time, in Fig.~\ref{fig:1}).
Our contributions are as follows:
\begin{itemize}

\item We propose MF-NeRF, an efficient NeRF framework that aims to reduce parameter size and training time while keeping the reconstruction quality high. We first design a \textit{mixed-feature hash encoding} to adaptively mix partial feature grids into a single hash table. To ensure accurate feature vector retrieval from this table, we further design an \textit{index transformation} technique that standardizes grid indices from distinct grid spaces to a shared grid space.

    \item We conduct extensive experiments to benchmark MF-NeRF against state-of-the-art (SOTA) methods such as Instant-NGP, TensoRF, and DVGO. Our MF-NeRF method achieves the fastest training time on the same GPU hardware while maintaining similar or higher rendering quality. For example, compared to Instant-NGP, our MF-NeRF-8 ($2^{22}$) setup achieves around 10\% training time reduction on the same GPU with $0.15$ higher average PSNR on the synthetic NeRF dataset. 
\end{itemize}


\section{Related work}
\subsection{Neural Scene Representations}
Inspired by NeRF \cite{Mildenhall2020NeRF}, many works \cite{Lin2021BARF,Barron2021MipNeRF,Hwang2023EvNeRF,YenChen2021iNeRF,Suhail2022Light,Orsingher2022Learning,Kurz2022AdaNeRF,Chen2022MobileNeRF} have been proposed for novel view synthesis tasks. Generally, NeRF \cite{Mildenhall2020NeRF} encodes a scene into a couple of MLP networks, using multiple posed images. Therefore, having consistency between images is crucial, as NeRF struggles to learn the scene when provided with inaccurate pose information or impaired images. Earlier works such as Deblur-NeRF \cite{Ma2022Deblur} and PDRF \cite{Peng2022PDRF} have 
investigated approaches to train a clear scene from blurry images. Deblur-NeRF optimizes a separate neural network that generates rays to simulate camera motion and defocus blurs, thereby enabling NeRF to capture and represent deblurred scenes. Similarly, in cases where image noise arises from low-light environments or overheated sensors, PDRF \cite{Mildenhall2022NeRFintheDark} approaches the problem by modeling the noise using color blending, preventing NeRF from learning it. On the other hand, NAN \cite{Pearl2022NAN} focuses on generating a clean image with the same view from a burst of noisy images that have relatively small camera movement, in contrast to what Deblur-NeRF and PDRF assume. In addition, Ev-NeRF \cite{Hwang2023EvNeRF} demonstrated NeRF is trainable with event camera data rather than regular posed RGB images.

Methods that leverage scene prior information continue to be an active area of research \cite{Lee2023DenseDepth,Orsingher2022Learning}. NerfingMVS \cite{Wei2021NerfingMVS} trains a neural network that completes sparse point clouds generated by Structure-from-Motion \cite{Schonberger2016Structure} into dense point clouds to help NeRF better understand the scene with the geometry. DS-NeRF \cite{Deng2022DepthSupervised} uses sparse point clouds to help learn the density distribution of the scene. MVG-NeRF \cite{Orsingher2022Learning} utilizes depth maps generated by multi-view stereo \cite{Schonberger2016Pixelwise} and compares them with synthesized depth maps produced by neural rendering methods. On the other hand, iNeRF \cite{YenChen2021iNeRF} propose a localization method that estimates the poses of given images by comparing them with neural-rendered images based on possible poses.

NeRF updates the entire MLP network for a given image set, which means it can learn only one scene at a time. Recent research \cite{Wang2021IBRNet,Wang2022Generalizable,Johari2022GeoNeRF} has proved that transformers \cite{Vaswani2017Attention} can be employed to select images from a given set for synthesizing novel view images, rather than directly encoding a scene into MLPs. Additionally, considering epipolar geometry constraints can enhance the restoration of view-dependent lighting effects \cite{Suhail2022Light}.

\subsection{Efficient NeRF Rendering Algorithms}
To address the slow training and long rendering time of NeRF, recent research has begun to use explicit data structures for storing and accessing features that encode the shape and color information of a scene \cite{Liu2020Neural,Martel2021Acorn,Hedman2021Baking}. NSVF \cite{Liu2020Neural} employs an Octree to store optimized scene features. This approach supports the gradual subdivision of voxels and the pruning of empty cells, which allows increased scene resolution and improved rendering quality. DVGO \cite{Sun2022Direct} adaptively increases the resolution of the feature grid during training. In contrast, KiloNeRF \cite{Reiser2021KiloNeRF} uses numerous amount of tiny MLPs instead of one large MLP to dramatically reduce rendering time. SNeRG \cite{Hedman2021Baking} extracts view-dependent and color features from a trained NeRF, bakes them into a 3D texture atlas, and uses a relatively small MLP to synthesize novel views with the baked features. RT-NeRF \cite{Li2022RTNeRF} performs fast rendering by back projecting the feature grid onto the render target image plane instead of using ray marching. Control-NeRF \cite{Lazova2023ControlNeRF} blends feature grids trained on different scenes to selectively render objects across the scenes.

A recent seminal work, Instant-NGP \cite{mueller2022instant}, maps multiresolution grids that encode a scene at various levels of abstraction onto a relatively small memory space using a hash function, allowing for the rapid learning of a plausible scene in seconds. In addition, TensoRF \cite{Chen2022TensoRF} decomposes a deep feature grid into vectors and matrices, resulting in improved storage efficiency.



\section{Method}


\begin{figure*}
    \centering
    \includegraphics[width=\textwidth]{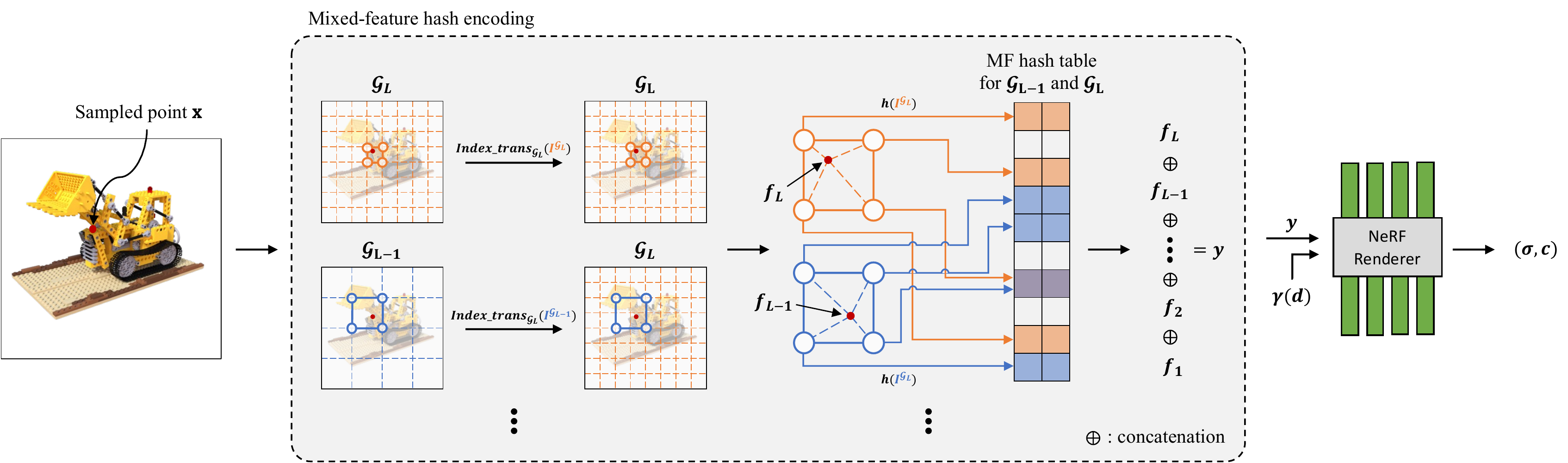}
    \caption{Overview of MF-NeRF in 2D representation. MF-NeRF employs mixed-feature hash encoding to share scene features across consecutive grids, allowing for mixing fine and coarse details. Indices from lower grids are remapped to the highest resolution grid sharing the same MF hash table, allowing grids to reference identical hash entries for congruent scene coordinates, as exemplified by the lower-right corners of the orange and blue voxels in the figure. After interpolating feature vectors of the corners, the NeRF Renderer processes these along with the positional embedded viewing direction $\gamma(\mathbf{d})$ to determine point $\textbf{x}$'s density $\sigma$ and color $\mathbf{c}$. Our method could be simply extended to three dimensions as well.
    }
    \label{fig:2}
\end{figure*}

\subsection{Preliminary}
\label{subsec:background}

\paragraph{Scene representation of neural radiance field}
NeRF~\cite{Mildenhall2020NeRF} uses MLP networks to learn the radiance field of arbitrary scenes. Once sufficiently optimized, NeRF can estimate both the volume density $\mathbf{\sigma}$ and RGB color $\mathbf{c}$ of the continuous function that describes the radiance field. It is then able to reconstruct a photo-realistic view from an unseen pose using a differentiable volume rendering method.

Given a camera pose and pixel coordinates, NeRF generates a camera ray $\mathbf{r}(t)=\mathbf{o}+t\mathbf{d}$ that marches from $\mathbf{o}\in\mathbb{R}^3$ in direction $\mathbf{d}\in[-\pi,\pi]^2$ to the pixel and samples $N$ points along the ray $\mathbf{r}$ with a certain distribution. Each sampled point $\mathbf{x}\in\mathbb{R}^3$ is given as input to the MLP $F_\Phi: (\mathbf{x}, \mathbf{d}) \mapsto (\sigma, \mathbf{c})$. 
The MLP further infers the density $\sigma\in\mathbb{R}$ of the point $\mathbf{x}$ and  RGB color $\mathbf{c}\in\mathbb{R}^3$ corresponding to the ray direction $\mathbf{d}$. 
Taking into account the density $\sigma_i$ and light transmittance $T_i$ of each point $\mathbf {x}_i$,
the final color $\hat{C}(\mathbf{r})$ for the ray $\mathbf{r}$ (i.e., the color of the pixel) is determined by a weighted sum of the colors of the $N$ sampled points along the ray $\mathbf{r}$, which can be formulated as: 
\begin{equation}
\begin{split}
\hat{C}(\mathbf{r}) &= \sum_{i=1}^{N} T_i (1-\text{exp}(-\sigma_i \delta_i)) \mathbf{c}_i, \\
T_i &= \prod_{j<i} \text{exp}(-\sigma_j \delta_j),
\end{split}
\end{equation}
where $\delta_i$ is the distance between successive sampled points $\text{x}_i$ and $\text{x}_{i+1}$, and $i$ and $j$ indicate the order of the sampled points.

\paragraph{Multiresolution hash encoding} To enhance training efficiency without compromising reconstruction quality, prior NeRF methods \cite{Sun2022Direct,Martel2021Acorn,Liu2020Neural,mueller2022instant} opt to encode the radiance field using feature grids with small MLP network instead of training a large MLP on the entire scene. 

Instant-NGP \cite{mueller2022instant}, as a representative work of this method, compresses multiple feature grids with different resolutions into dedicated hash tables for expressing smoothness and minimizing worthless grid points. 
Specifically, it owns two types of trainable parameters: encoding parameters $\Theta$ stored in the hash tables and weight parameters $\Phi$ for the NeRF Renderer. The encoding parameters are arranged into $L$ hierarchical grid levels with distinct resolutions. Each grid point contains a feature vector with an $F$-dimensional size. Furthermore, every feature vector for a given grid resides in its respective $l\text{-}th$ hash table, which has a maximum capacity of $T$. The resolution ratio $b$ between successive grids is defined as follows:
\begin{equation}
\label{eqt:ratio}
    b=(N_{max} / N_{min})^{1/(L-1)},
\end{equation}
where $N_{max}$ and $N_{min}$ are the maximum resolution and minimum resolution of grids, respectively. Hence, the resolution of $l\text{-}th$ grid is given by:
\begin{equation}
\label{eqt:nl}
   N_l = \lfloor N_{min} \cdot b^{l-1} \rfloor
\end{equation}

Given a sampled point $\textbf{x}$ and a viewing direction $\textbf{d}$, Instant-NGP identifies the indices, $\text{I}^{\mathcal{G}_l}=(i,j,k)$, of the eight voxel corners surrounding $\textbf{x}$ at each grids $\mathcal{G}_l$, resulting in the total of $8\cdot L$ corner points. Then, the spatial hash $h(\cdot)$ \cite{Teschner2003Optimized} is applied to each $\text{I}^{\mathcal{G}_l}$ to retrive the associated feature vector from the dedicated hash table $\theta_l$:
\begin{equation} \label{eq:2}
    h(\text{I}^{\mathcal{G}_l}) = (i \cdot \pi_1 \oplus j \cdot \pi_2 \oplus k \cdot \pi_3) \mod T,
\end{equation}
where $\oplus$ is bit-wise $\text{XOR}$ operator, $T$ is the hash table size, and $\pi$ is a prime number (i.e., $\pi_1 = 1$, $\pi_2 = 2,654,435,761$, and $\pi_3 = 805,459,861$).
After that, the eight $F$-dimensional feature vectors are trilinearly interpolated resulting in $f_{l} \in \mathbb{R}^F$. Then, the $L$ feature vectors are concatenated into $\textbf{y}= [ f_{1}; ...; f_{L} ]$, which serves as the input to the NeRF Renderer, with viewing direction $\textbf{d}$. The NeRF Renderer is constructed by a density MLP $m_d$ and a color MLP $m_c$ which are parameterized by $\Phi_d$ and $\Phi_c$ (i.e., $\Phi=[\Phi_d;\Phi_c]$), respectively. The density $\sigma$ and color $\textbf{c}$ estimation can be formulated like:
\begin{equation} \label{eq:mlps}
\begin{split}
    f_c   &= m_d(\textbf{y};\Phi_d),    \\
    \sigma  &= f_c[0],  \\
    \textbf{c}  &= m_c(f_c;\Phi_c), 
\end{split}
\end{equation}
where $f_c$ represents the color features produced by the density MLP $m_d$.

\paragraph{Limitations of multiresolution hash encoding}
However, although the  multiresolution hash encoding enables fast learning speed compared to conventional NeRF~\cite{Mildenhall2020NeRF}, there still remain several limitations. First, it inefficiently uses multiple dense feature grids to represent vast empty spaces and occupied spaces that inherently exist in 3D space. Due to the uniformity of the object's interior, incident voxels within the object across different grids are prone to containing similar feature values. This also applies to the empty spaces, resulting in many voxels across the grids having similar feature values. Second, multiresolution hash encoding compresses features into multiple hash tables corresponding to feature grids of different resolutions via hash function, 
which means it has to manage the same number of feature grids with the feature resolution levels $L$.
For example, coarse grid learns simple shapes and fine grid learns detailed shapes. 
Moreover, the features of each grid are saved in the different memory space, and the size of hash table $T$ and feature resolution levels $L$
crucially affect the learning capacity of Instant-NGP. This indicates that
higher values of $T$ and $L$ are essential for effectively reconstructing high-quality scenes with more intricate details. 
However, this results in substantial encoding parameters,
which is orders of magnitude larger than the weight parameters of the MLP. For instance, in a general setting of Instant-NGP with a hash table size $T$ of $2^{19}$ and feature resolution levels $L$ of 16, the resulting encoding parameter size is 12.6 M, which is \textbf{1,191$\times$} larger than the weight parameter size (10k). The extensive memory usage due to large encoding parameters lead to a notable increase in computer cache miss ratio, consequently resulting in significantly longer training time, especially when $T$ is larger than the size of computer cache~\cite{mueller2022instant}.
Likewise, this limitation could potentially restrict its practical implementation in real-world scenarios, such as large-scale view synthesis or deployment on memory-limited edge devices.



\subsection{Proposed MF-NeRF}
In this section, we present MF-NeRF, a memory efficient NeRF framework with a novel \textbf{mixed-feature hash table} design to reduce   
memory usage and training time, all while preserving the quality of reconstruction as shown in Fig.~\ref{fig:2}.
Unlike Instant-NGP which learns different levels of abstraction of the scene with dedicated grid size (namely, the number of grid 
with different sizes equal to the different resolution levels), the proposed mixed-feature hash encodes mix multiple grids into a shared hash table, enabling the learning of combined coarse and fine abstracted features. 
This design choice is based on the assumption that overlapping grids in the same radiance field likely exhibit analogous feature distributions. Thus, maintaining separate hash tables for each grid becomes redundant. Instead, we allow similar features from different grids to be stored in the same entry of a shared mixed-feature hash table. 

Furthermore, to retrieve appropriate features from the MF hash table, we further design an \textbf{index transformation} method, which converts an index from a lower-level grid to a corresponding index in a higher-level grid linked via the same MF hash table.
For each of the $L$ grids, the feature vectors corresponding to the eight voxel corners are retrieved and trilinearly interpolated. Then, the $L$ interpolated feature vectors are concatenated and fed into a series of MLPs to estimate the density and color values of the sample point. The detailed techniques will be introduced in the following sections.

\subsubsection{Mixed-Feature Hash Encoding} \label{subsec:mixedfeature_hash}

First of all, following the setting in the Instant-NGP~\cite{mueller2022instant}, we construct $L$ grids $\{\mathcal{G}_l\}_{l=1}^{L}$ with different resolution levels. The resolution ratio $b$ between grid $\mathcal{G}_l$ and $\mathcal{G}_{l+1}$ aligns with definition as shown in Eq.~\ref{eqt:ratio}. 

To build mixed-feature hash table, we create $N \in [1, L]$ hash tables, each with a maximum size of $T \in [2^{14}, 2^{24}]$. In practice, the size of hash table size is given by $min(T, N_{i \cdot W}^3)$, where $i \in [1,N]$, and each entry holds an $F$-dimensional feature vector. The selection of hyperparameter $N$ should be less than or equal to the number of grids $L$. For example, if $N$ equals to $L$, the number of hash table is same as the multiresolution hash encoding; When $N=1$, a single hash table is shared by all grids. 

Furthermore, to mix multiple grids with different levels into one hash table, we develop an index transformation method to transform the current index space of these grids to the index space of the finest grid for each hash table with the corresponding grid groups (The detailed technique of index transformation will be introduced below). For example, given that $L=8$ and $N=2$, each 4 consecutive grids are encoded into one hash table. More specifically, when a sampled point $\textbf{x}$ is queried, the indices of the eight voxel corner points surrounding the point $\textbf{x}$ at each grids $\{\mathcal{G}_l\}_{l=1}^{8}$ will be first identified. Following that, the indices $\{\text{I}^{\mathcal{G}_l}\}_{l=1}^{4}$ from the initial half of the grids are transformed to the corresponding indices $\text{I}^{\mathcal{G}_4}$ of the 4th-level grid $\mathcal{G}_4$, while the indices $\{\text{I}^{\mathcal{G}_l}\}_{l=5}^{8}$ from the latter half are transformed to the corresponding indices $\text{I}^{\mathcal{G}_8}$ of the 8th-level grid $\mathcal{G}_8$. 




Upon obtaining the transformed indices, they are hashed to find the entries in the defined hash table by using the spatial hash function as shown in Eq.~\ref{eq:2}. The resultant eight $F$-dimensional feature vectors undergo trilinear interpolation to produce $\{f_l\}_{l=1}^{L}$. These $L$ feature vectors are concatenated as $\textbf{y}= [ f_{1}; ...; f_{L} ]$ and then fed into the NeRF Renderer (Eq.~\ref{eq:mlps}) alongside the viewing direction $\textbf{d}$, yielding the density and color values of $\textbf{x}$.
By doing so, The mixed-feature (MF) hash table is built to  consolidates partial grids within a single hash table. It allows the proposed MF-NeRF to learn mixed features of coarse and fine abstraction, leading to significantly feature size reduction.

\subsubsection{Index Transformation} \label{subsec:window_to_grid}
Given that voxel size varies depending on grid resolution, grid points at identical positions in different grids might have distinct indices. For example, in Fig.~\ref{fig:2}, both the orange and blue voxels' lower-right corners are located at the same position within the lego scene. However, their indices are $(2,2)$ in grid $\mathcal{G}_{L-1}$ and $(4,4)$ in grid $\mathcal{G}_{L}$. When hashed, these indices map to different entries in the MF hash table despite representing the same spatial location. To address this, we employ an index transform function $\textit{Index\_trans}_{\mathcal{G}_j}(\cdot)$ to translate indices in arbitrary level grid $\mathcal{G}_{i}$ into those in $j\text{-}th$ level grid $\mathcal{G}_{j}$ prior to hashing.
\begin{equation}
\begin{split}
\textit{Index\_trans}_{\mathcal{G}_j}(\text{I}^{\mathcal{G}_i}) &= \lfloor \text{I}^{\mathcal{G}_i} \times N_{j} / N_{i} \rfloor \\
                                    &= \text{I}^{\mathcal{G}_j},
\end{split}
\label{eqt:index}
\end{equation}
where $N_{\mathcal{G}_i}$ and $N_{\mathcal{G}_j}$ are resolutions of $i\text{-}th$ and $j\text{-}th$ grids calculated by Eq.~\ref{eqt:nl}. With this function, we obtain the indices in the finest grid, among those sharing the same MF hash table, for grid points of grids at any levels. Using consistent grid space indices when accessing the MF hash table ensures that multiple grids reference the same feature for a specific location.



\subsubsection{Training} \label{subsec:training}
The proposed MF-NeRF is trained by using a set of posed images. 
Specifically, we jointly optimize both encoding and weight parameters by calculating the $\mathcal{L}^2$ loss between the rendered color $\hat{C}(\textbf{r})$ and the ground truth color $C(\textbf{r})$. The loss function is defined as:
\begin{equation}
\textit{Loss} = \sum_{\textbf{r} \in \mathcal{R}} \mid \mid \hat{C}(\textbf{r}) - C(\textbf{r}) \mid \mid^2_2
\end{equation}
where $\mathcal{R}$ is a batch of camera rays randomly sampled from all the pixels in the image set.

In summary, the whole training process can be illustrated as follows: given a sampled point $\textbf{x}$ and a viewing direction $\textbf{d}$, MF-NeRF first obtains the voxel's eight corner indices $\{\text{I}^{\mathcal{G}_l}\}_{l=1}^{L}$ at each grids. Before hashing the eight ${\text{I}^{\mathcal{G}_l}}_{l=1}^{L}$, these indices are transformed to the corresponding highest-resolution grid space by using the index transformation function.
After this transformation, corresponding feature vectors are retrieved from the MF hash tables. The eight feature vectors corresponding to the voxel's eight corner points are then trilinearly interpolated, resulting in one feature vector $f_{l}$ for $l\text{-}th$ level grid. Then, $L$ interpolated feature vectors are concatenated (i.e., $\textbf{y}=[f_{1};...;f_{L}]$) and forwarded to the NeRF Renderer along with the viewing direction $\textbf{d}$ (Eq.~\ref{eq:mlps}).

\section{Experiments}

\begin{table}[t]
\centering
\begin{tabular}{@{}rcrc@{}}
\toprule
\multicolumn{2}{c}{Instant-NGP}                                     & \multicolumn{2}{c}{MF-NeRF-1}                  \\
\multicolumn{1}{c}{\# Features (T)} & PSNR                       & \multicolumn{1}{c}{\# Features (T)} & PSNR  \\ \midrule
3,293,600 ($2^{17}$)      & \multicolumn{1}{c|}{32.69} & 2,097,152 ($2^{20}$)      & 32.70 \\
6,177,184 ($2^{18}$)      & \multicolumn{1}{c|}{32.76} & 4,194,304 ($2^{21}$)      & 32.91 \\
11,445,040 ($2^{19}$) & \multicolumn{1}{c|}{32.91} & 8,388,608 ($2^{22}$)  & 33.10 \\
21,061,904 ($2^{20}$) & \multicolumn{1}{c|}{32.97} & 16,777,216 ($2^{23}$) & 33.13 \\ \bottomrule
\end{tabular}
\caption{MF-NeRF with one MF hash table ({MF-NeRF-1}) can learn a scene with smaller memory usage.}
\label{tab:1}
\end{table}

\begin{table*}[!t]
\centering
\setlength{\tabcolsep}{1pt}
\footnotesize
\scalebox{0.95}{
\begin{tabular}{@{}l|rrrr|cccc|cccc|cccc|cccc@{}}
\toprule
\multirow{2}{*}{\begin{tabular}[c]{@{}l@{}}Synthetic\\ NeRF\end{tabular}} &
  \multicolumn{4}{c|}{\# Features} &
  \multicolumn{4}{c|}{PSNR} &
  \multicolumn{4}{c|}{SSIM} &
  \multicolumn{4}{c|}{LPIPS} &
  \multicolumn{4}{c}{Training Time (mins)} \\ \cmidrule(l){2-21} 
 &
  \multicolumn{1}{c}{$2^{20}$} &
  \multicolumn{1}{c}{$2^{21}$} &
  \multicolumn{1}{c}{$2^{22}$} &
  \multicolumn{1}{c|}{$2^{23}$} &
  $2^{20}$ &
  $2^{21}$ &
  $2^{22}$ &
  $2^{23}$ &
  $2^{20}$ &
  $2^{21}$ &
  $2^{22}$ &
  $2^{23}$ &
  $2^{20}$ &
  $2^{21}$ &
  $2^{22}$ &
  $2^{23}$ &
  $2^{20}$ &
  $2^{21}$ &
  $2^{22}$ &
  $2^{23}$ \\ \midrule
Instant-NGP &
  21.06 M &
  38.55 M &
  70.20 M &
  126.97 M &
  32.97 &
  33.03 &
  33.05 &
  33.00 &
  .9595 &
  .9602 &
  .9602 &
  .9599 &
  .0501 &
  .0490 &
  .0482 &
  .0478 &
  8.42 &
  10.93 &
  13.55 &
  17.43 \\
MF-NeRF-1 &
  2.10 M &
  4.19 M &
  8.39 M &
  16.78 M &
  32.70 &
  32.91 &
  33.10 &
  33.13 &
  .9572 &
  .9582 &
  .9603 &
  .9597 &
  .0560 &
  .0533 &
  .0505 &
  .0496 &
  8.17 &
  9.84 &
  11.27 &
  12.49 \\
MF-NeRF-2 &
  4.19 M &
  7.00 M &
  11.20 M &
  19.59 M &
  32.80 &
  32.88 &
  33.07 &
  33.04 &
  .9578 &
  .9595 &
  .9601 &
  .9602 &
  .0544 &
  .0519 &
  .0502 &
  .0494 &
  8.23 &
  9.82 &
  11.36 &
  12.55 \\
MF-NeRF-4 &
  6.39 M &
  11.30 M &
  19.69 M &
  36.47 M &
  32.91 &
  33.05 &
  33.08 &
  33.08 &
  .9594 &
  .9600 &
  .9601 &
  .9604 &
  .0519 &
  .0497 &
  .0490 &
  .0486 &
  8.33 &
  9.92 &
  11.38 &
  12.96 \\
MF-NeRF-8 &
  11.16 M &
  20.26 M &
  37.04 M &
  68.64 M &
  33.06 &
  33.11 &
  33.20 &
  33.15 &
  .9600 &
  .9610 &
  .9606 &
  .9611 &
  .0499 &
  .0489 &
  .0480 &
  .0473 &
  8.37 &
  10.34 &
  12.18 &
  14.58 \\ \bottomrule
\end{tabular}}
\caption{Ablation study on different number of mixed-feature tables and table sizes. The trailing numbers of MF-NeRF- indicates the number of MF hash tables $N$.}
\label{tab:2}
\end{table*}

\begin{table}[!t]
\centering
\setlength{\tabcolsep}{2pt}
\scalebox{0.8}{
\begin{tabular}{@{}l|l|c|cccc@{}}
\toprule
Dataset                         & Method                              & \# Features & PSNR  & SSIM  & LPIPS & Time  \\ \midrule
\multirow{7}{*}{\begin{tabular}[c]{@{}l@{}}Synthetic\\ NeRF\end{tabular}} & TensoRF                             & 17.34 M     & 33.22 & .9630 & .0470 & 28.77 \\
                                & DVGOv1                              & 57.34 M     & 31.94 & .9567 & .0534 & 19.02 \\
                                & DVGOv2                              & 57.34 M     & 31.90 & .9561 & .0539 & 8.58  \\
                                & Instant-NGP ($2^{20}$) & 21.06 M     & 32.97 & .9595 & .0501 & 8.42  \\
                                & Instant-NGP ($2^{22}$) & 70.20 M     & 33.05 & .9602 & .0482 & 13.55 \\
                                & MF-NeRF-8 ($2^{20}$)   & 11.16 M     & 33.06 & .9600 & .0499 & 8.37  \\
                                & MF-NeRF-8 ($2^{22}$)   & 37.03 M     & 33.20 & .9606 & .0480 & 12.18 \\ \midrule
\multirow{5}{*}{MipNeRF360}     & DVGOv2                              & 57.34 M     & 25.70 & .7043 & .3654 & 24.41 \\
                                & Instant-NGP ($2^{20}$) & 21.06 M     & 27.06 & .7455 & .3243 & 12.44 \\
                                & Instant-NGP ($2^{22}$) & 70.20 M     & 27.36 & .7699 & .2939 & 22.81 \\
                                & MF-NeRF-8 ($2^{20}$)   & 11.16 M     & 27.11 & .7400 & .3342 & 11.78 \\
                                & MF-NeRF-8 ($2^{22}$)   & 37.03 M     & 27.38 & .7641 & .3016 & 20.31 \\ \bottomrule
\end{tabular}
}
\caption{Quantitative comparison with SOTA works (the numbers in the parentheses of methods indicate the hash table size $T$)}
\label{tab:3}
\end{table}
\subsection{Experimental setup}

\paragraph{Dataset}
We train MF-NeRF and the baselines on the widely-use Synthetic NeRF dataset \cite{Mildenhall2020NeRF}, which includes eight different synthetic scenes. Each scene contains $400$ posed images: $100$ for training, $100$ for validation, and $200$ for testing, all with a resolution of $800\times800$. Additionally, we train on another popular MipNeRF360 dataset \cite{Barron2022MipNeRF}, which encompasses seven publicized, unbounded real scenes. Each scene consists of between $125$ and $311$ images. Since this dataset is not pre-split, we select every 8th image for testing. Image resolutions vary by scene, ranging from $3114\times2075$ to $4978\times3300$.

\paragraph{Metric}
We employ three commonly used metrics to evaluate the performance: peak signal-to-noise ratio (PSNR), structural similarity index measure (SSIM) \cite{Wang2004Image}, and learned perceptual image patch similarity of VGG (LPIPS-VGG) \cite{Zhang2018Unreasonable}. These metrics quantify the similarity between the ground truth images and the reconstructed images.

\paragraph{Baselines:}
We consider Instant-NGP~\cite{mueller2022instant} as our baseline which uses a multiresolution hash table to encode scene features. Moreover, we also compare to other two SOTA methods: TensoRF~\cite{Chen2022TensoRF}, DVGOv1~\cite{Sun2022Direct}, and DVGOv2~\cite{Sun2022Improved}.

\paragraph{Implementation details}

We adopt Adam \cite{Kingma2014Adam} optimizer with an initial learning rate of $2 \cdot 10^{-2}$, which decreased to $2 \cdot 10^{-4}$ using cosine decay. The random seed is set to 1337. 
All the synthetic scenes are trained with the same hyperparameters, including a batch size of $16384$, $N_{max}=1024$, $N_{min}=16$, $L=16$, and $F=2$. For the unbounded scenes, we train MF-NeRF with hyperparameters including a batch size of $4096$, $N_{max}=1024$, $N_{min}=16$, $L=16$, and $F=2$. We use $0.25$ times downsampled images for the unbounded scene training. We configure different scale factors for the unbounded scenes ($16\times$ for bicycle, bonsai, counter, and garden scenes; $4\times$ for kitchen and room scenes; $64\times$ for stump scene). We train $20$k iterations for both synthetic and unbounded scenes on a single NVIDIA Quadro RTX 5000 GPU.

In our experiments, the density MLP is configured with a single hidden layer with 64 channels, while the color MLP consists of two hidden layers, each with 128 channels. Empirical testing led us to choose broader channels for the color MLP compared to Instant-NGP, maximizing the MLP's performance. This optimized MLP offers improved capability in both mitigating hash collisions and translating color features into more realistic colors. Unless otherwise specified, we set $N=8$ for MF-NeRF as the default configuration.


\subsection{Experimental Results}
\paragraph{Scene compression}
First, we investigate how MF-NeRF can effectively compress a scene. To do so, we train MF-NeRF in the extreme case where there is only one single MF hash table (i.e., $N=1$). To fair compare to Instant-NGP, we only adjust the hash table size, while keeping the other settings same.
In Table~\ref{tab:1}, we show the results of training Instant-NGP and MF-NeRF-1 ($N=1$, $W=16$) on the Synthetic NeRF dataset (both methods have $L=16$, $F=2$, $N_{min}=16$, $N_{max}=1024$, $\text{batch}=16384$, $\text{iters}=20$k settings in common). 
It is intriguing to observe that
MF-NeRF-1 achieves better performance than Instant-NGP with 20-36\% fewer parameters. We believe this result suggests that Instant-NGP may suffers from unnecessary redundancy due to 
learn similar features among each grid.


\paragraph{Comparison with Instant-NGP}
While sharing a single hash table among all grids (i.e., MF-NeRF-1) can significantly reduce memory usage, it may also degrade learning performance due to excessively limited memory space. 
To determine the ideal value for $N$, we conduct experiments by doubling the number of MF hash tables from $N=1$ up to $N=8$.
As shown in Table~\ref{tab:2}, we report the performance of MF-NeRF-$N$ with different $N$ settings of $1$, $2$, $4$, and $8$. Compared to Instant-NGP, we can see that MF-NeRF-8 achieved a parameter reduction of $47.01\%$/$47.44\%$/$47.24\%$/$45.94\%$, and training time reductions of $3s$/$35s$/$82s$/$171s$, with a $0.09$/$0.08$/$0.15$/$0.15$ PSNR improvement at all four different $T$ settings, respectively. In addition, it is interesting to see that as the hyperparameter $T$ increased, MF-NeRF showed larger training time reductions compared to Instant-NGP, indicating that the larger parameters of Instant-NGP are a bottleneck that constrains its training time. 



\begin{figure*}[!t]
    \centering
    \includegraphics[width=0.85\textwidth]{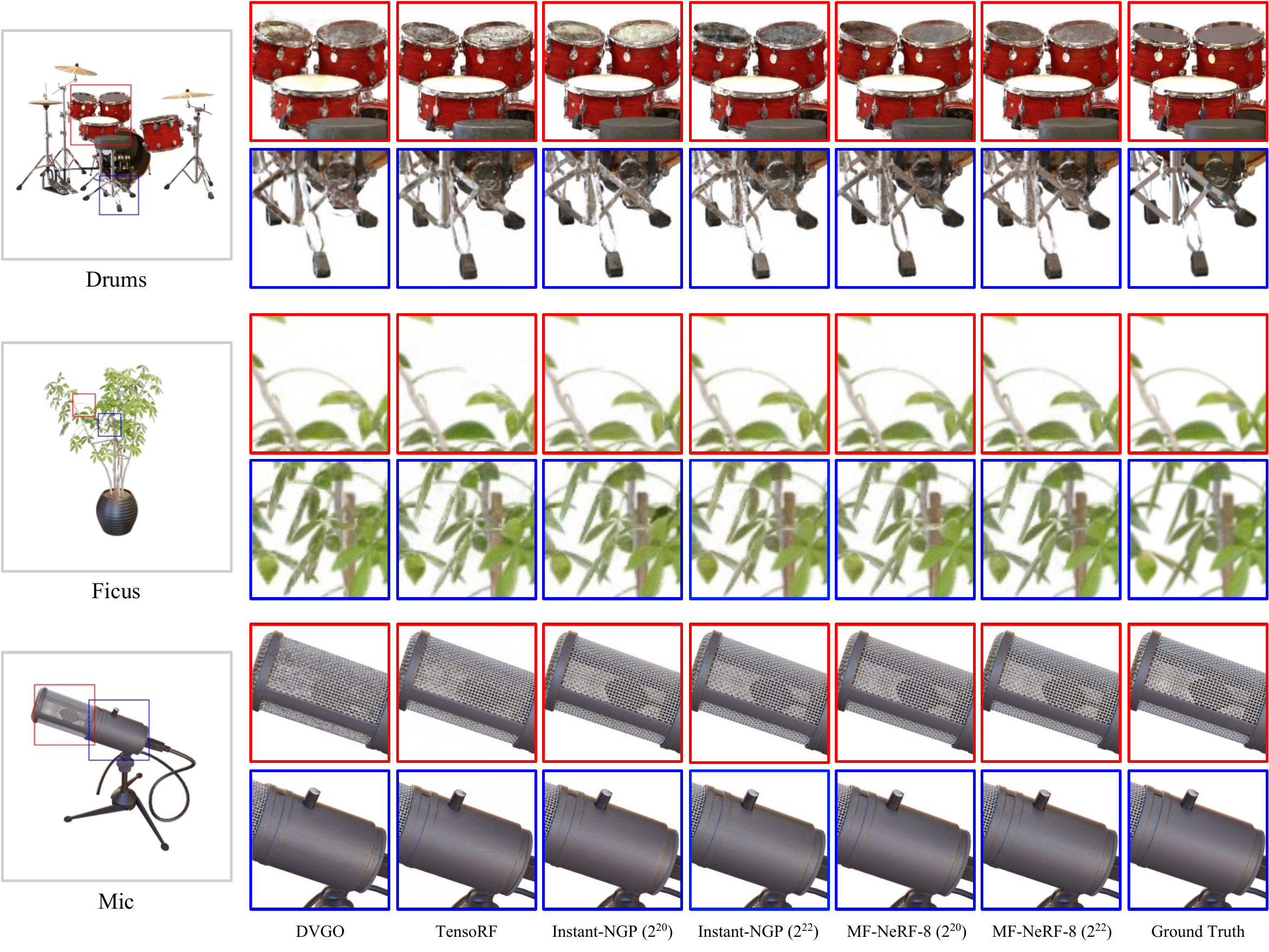}
    \caption{We present qualitative results on three scenes, comparing MF-NeRF-8 with other SOTA methods.}
    \label{fig:3}
\end{figure*}

\paragraph{Benchmarking with SOTA methods}
We further compare the performance of MF-NeRF with other SOTA methods, including TensoRF \cite{Chen2022TensoRF}, DVGOv1 \cite{Sun2022Direct}, DVGOv2~\cite{Sun2022Improved}, and Instant-NGP \cite{mueller2022instant} using the Synthetic NeRF and MipNeRF360 datasets.
The comprehensive comparison results are reported in Table~\ref{tab:3}. For the Synthetic NeRF benchmark, although TensoRF shows overall high performance, it has the largest training time, with around $3.4\times$ and $2.3\times$ longer training time than our MF-NeRF-8 ($2^{20}$) and MF-NeRF-8 ($2^{22}$), respectively. However, the average PSNR difference is only $0.16$ and $0.02$, respectively.
As can be seen, both MF-NeRF-8 ($2^{20}$) and MF-NeRF-8 ($2^{22}$) configurations learn the scenes faster and exhibit higher average PSNR and SSIM than DVGOv1. When considering the same $T$, MF-NeRF-8 consistently outperforms its Instant-NGP counterparts in terms of average PSNR and requires a shorter training time.
Compared to DVGOv2, our MF-NeRF-8 ($2^{20}$) achieves better average PSNR and shorter training time on all synthetic scenes. Specifically, we achieve a $1.16$ PSNR improvement and a slight $12s$ training time reduction on average. In addition, compared to TensoRF, our MF-NeRF-8 ($2^{22}$) achieves almost the same PSNR (i.e., $0.01$ degradation) but significantly reduces the training time by up to $2.4\times$ on average. 

In the MipNeRF360 benchmark tests, our models MF-NeRF-8 ($2^{20}$) and MF-NeRF-8 ($2^{22}$) surpass DVGOv2, exhibiting average PSNR improvements of $1.41$/$1.68$ respectively, alongside training time reductions of $51.74\%$/$16.79\%$. Moreover, in comparisons with Instant-NGP of $T$ counterpart, these models consistently showcase superior reconstruction, with PSNR enhancements of $0.05$/$0.02$, and training time savings of $39s$/$150s$, respectively.




\subsection{Visualization}
Fig.~\ref{fig:3} shows a qualitative comparison between NeRF methods on the test images. When comparing the drum surface and legs, it can be observed that MF-NeRF depicts smoother and more accurate shading than Instant-NGP. 
We can also observe a similar phenomenon in the leaves, which Instant-NGP oddly depicts a dark green color. When considering volume reconstruction, MF-NeRF appears to not miss thin structures. MF-NeRF also recognizes the microphone sensor inside the thin mesh cover in the Mic scene and can depict slender branches without disconnection in the Ficus scene.


\section{Conclusion}
In this paper, we propose a novel framework for neural radiance field (NeRF) that employs a mixed-feature (MF) hash table to improve memory efficiency, and reduce training/rendering time, while maintaining reconstruction quality. 
By sharing the hash table, our method saves memory space for storing scene features and enhances the training speed by preventing multiple feature grids from redundantly learning the same feature.
In the experiment, we observe that our proposed method achieves significant improvements in terms of training time reduction and rendering quality with smaller parameter size compared to the prior Instant-NGP method.




\clearpage

\bibliography{references}

\end{document}



\section{Appendix A. Experiment detail}\label{sec:head}

We present detailed experimental results for each individual scene. Our experiments is conducted on the Synthetic NeRF \cite{Mildenhall2020NeRF} and MipNeRF360 \cite{Barron2022MipNeRF} datasets. As mentioned in the main text, we maintain consistent hyperparameters across the datasets with the exceptions of the image downsampling ratio, batch size, and scene scaling factor. We will elaborate on these scene-specific hyperparameter settings in a subsequent section. All other parameters are held constant, as summarized in Table~\ref{tab:0}, which are used for MF-NeRF and Instant-NGP \cite{mueller2022instant}.

\begin{table}[ht]
\centering
\begin{tabular}{@{}l|l|l@{}}
\toprule
\begin{tabular}[c]{@{}l@{}}Hyper-\\ parameter\end{tabular} & Description                             & Value                                           \\ \midrule
F              & \begin{tabular}[c]{@{}l@{}}Number of features \\ per hash table entry\end{tabular} & 2                                               \\
L              & Number of feature grids                 & 16                                              \\
N\textsubscript{min}     & Coarsest resolution of feature grid     & 16                                              \\
N\textsubscript{max}     & Finest resolution of feature grids      & 1024                                            \\
lr             & Learning rate                           & 2$\cdot$10\textsuperscript{-2} \\
decay          & Learning rate decay factor              & 1$\cdot$10\textsuperscript{-2} \\
iters          & Number of algorithm steps               & 20k                                             \\
scale          & Scene scaling factor                    & 1                                               \\
seed           & Random seed                             & 1337                                            \\ \bottomrule
\end{tabular}
\caption{Common hyperparameters for MF-NeRF and Instant-NGP over the three datasets.}
\label{tab:0}
\end{table}

For DVGOv1 \cite{Sun2022Direct}, DVGOv2 \cite{Sun2022Improved}, and TensoRF \cite{Chen2022TensoRF}, we use their default hyperparameter settings. We built our method directly on the CUDA code of Instant-NGP. However, for easy benchmarking of datasets not supported by Instant-NGP, we utilized the python implementation of Instant-NGP named `ngp\_pl' available at \url{https://github.com/kwea123/ngp_pl}.

\subsubsection{Synthetic NeRF dataset} 
\begin{figure*}[t]
\centering
\includegraphics[width=\textwidth]{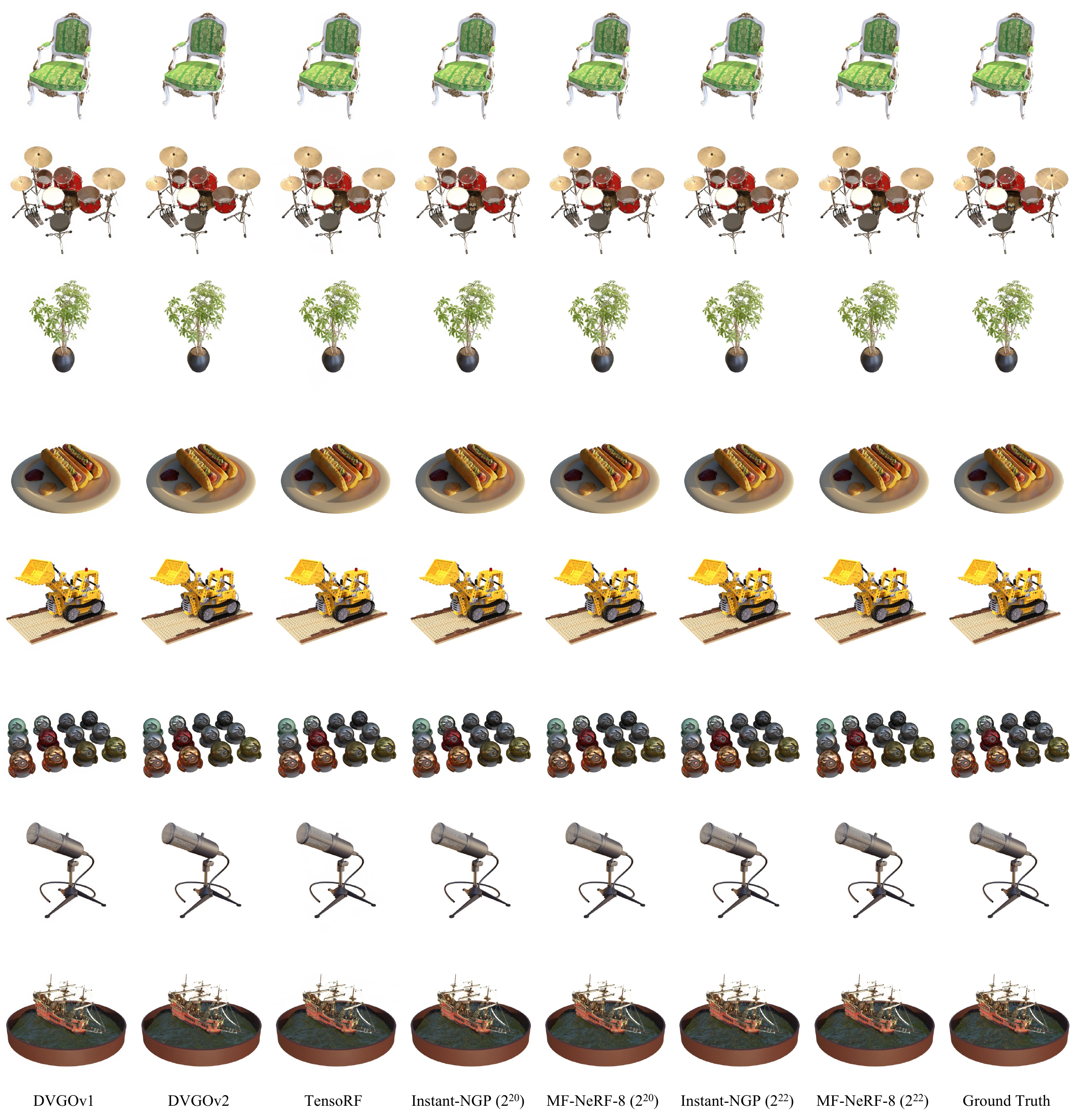}
\caption{Qualitative result on Synthetic NeRF dataset.}
\label{fig:1}
\end{figure*}

comprises eight scenes: Chair, Drums, Ficus, Hotdog, Lego, Materials, Mic, and Ship. Each scene provides 400 images with a resolution of $800\times800$. These images are divided as follows: 200 for training, 100 for evaluating, and 100 for testing. As the evaluation dataset is generally not utilized, we have omitted it, focusing solely on the training and testing datasets. We employed a consistent set of hyperparameters across all eight scenes for the MF-NeRF and Instant-NGP models: an image downsampling ratio of 1, a batch size of 16384, and a scene scaling factor of 1. The results are presented in Table~\ref{tab:1} and Fig.~\ref{fig:1}.

\subsubsection{MipNeRF360 dataset} 
\begin{figure*}[t]
\centering
\includegraphics[width=\textwidth]{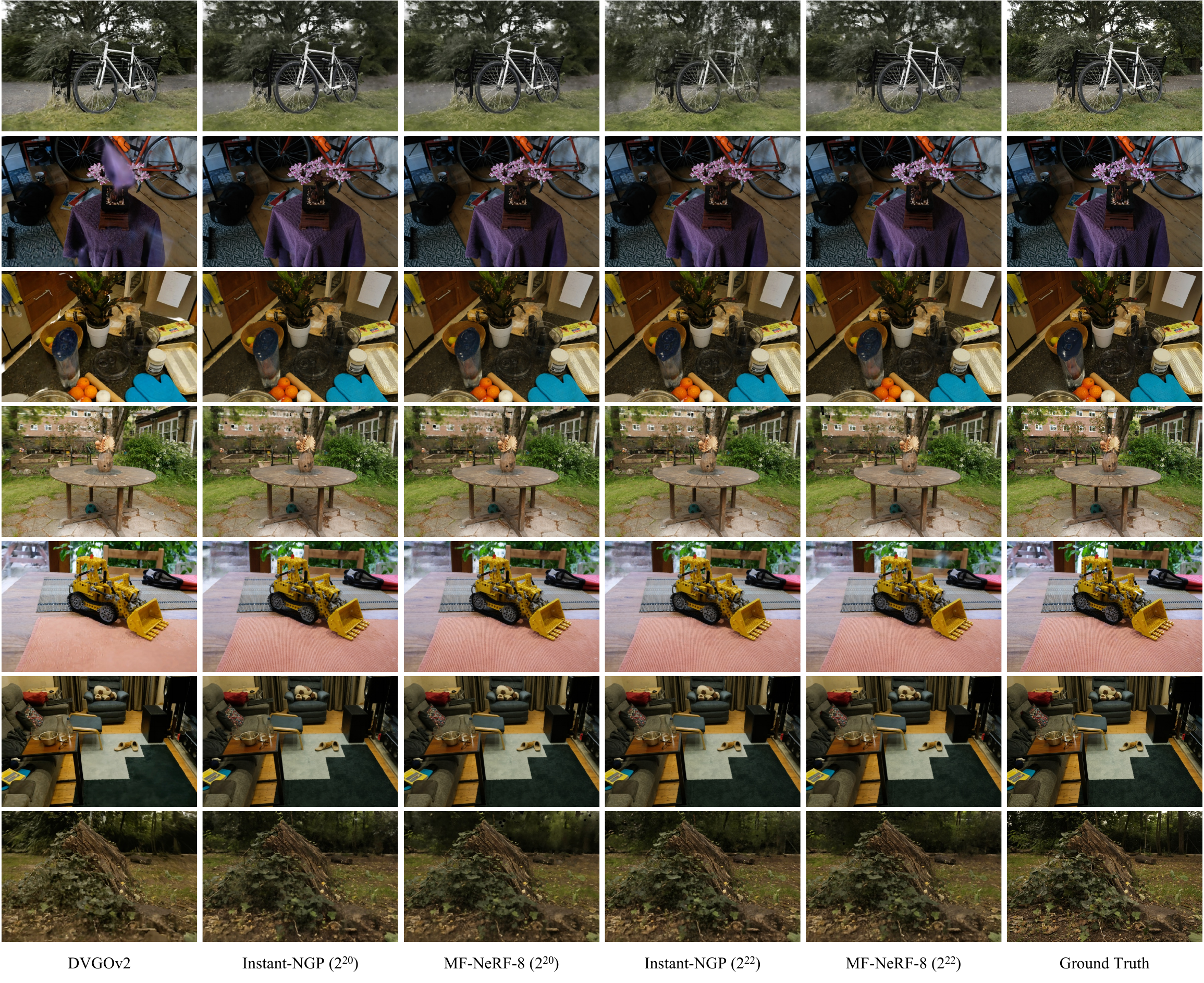}
\caption{Qualitative result on the MipNeRF360 dataset.}
\label{fig:2}
\end{figure*}

comprises seven scenes having backgrounds. These scenes are: Bicycle (194 images), Bonsai (292 images), Counter (240 images), Garden (185 images), Kitchen (279 images), Room (311 images), and Stump (125 images). We detail the properties and scale settings of these scenes in Table~\ref{tab:2-1}. For MF-NeRF and Instant-NGP model training, we employed a 1/4 downsampling ratio and used batches of 4096 rays. The results can be found in Table~\ref{tab:2} and Fig.~\ref{fig:2}.

\begin{table}[!ht]
\centering
\begin{tabular}{@{}l|c|c|c@{}}
\toprule
Scene & \begin{tabular}[c]{@{}c@{}}Original \\ resolution\end{tabular} & \begin{tabular}[c]{@{}c@{}}Train/Test\\ split images\end{tabular} & Scale \\ \midrule
Bicycle & $4946 \times 3286$ & 169/25           & 16    \\
Bonsai  & $3118 \times 2078$ & 255/37           & 16    \\
Counter & $3115 \times 2076$ & 210/30           & 16    \\
Garden  & $5187 \times 3361$ & 161/24           & 16    \\
Kitchen & $3115 \times 2078$ & 244/35           & 4     \\
Room    & $3114 \times 2075$ & 272/39           & 4     \\
Stump   & $4978 \times 3300$ & 109/16           & 64    \\ \bottomrule
\end{tabular}
\caption{Scene by scene properties and scale hyperparameter setting.}
\label{tab:2-1}
\end{table}

\begin{table*}[t]
\centering
\begin{tabular}{@{}c|l|cccccccc|c@{}}
\toprule
Metrics &
  \multicolumn{1}{c|}{Methods} &
  Chair &
  Drums &
  Ficus &
  Hotdog &
  Lego &
  Materials &
  Mic &
  Ship &
  Avg. \\ \midrule
\multirow{7}{*}{PSNR}  & TensoRF                & 35.77 & 25.93 & 34.12 & 37.58 & 36.59 & 30.11 & 34.95 & 30.71 & 33.22 \\
                       & DVGOv1                 & 34.08 & 25.46 & 32.78 & 36.70 & 34.56 & 29.58 & 33.20 & 29.13 & 31.94 \\
                       & DVGOv2                 & 34.06 & 25.40 & 32.56 & 36.75 & 34.67 & 29.58 & 33.18 & 29.02 & 31.90 \\
                       & Instant-NGP ($2^{20}$) & 35.47 & 25.69 & 33.84 & 37.47 & 35.88 & 29.49 & 35.63 & 30.32 & 32.97 \\
                       & Instant-NGP ($2^{22}$) & 35.49 & 25.79 & 34.03 & 37.41 & 36.00 & 29.53 & 35.88 & 30.30 & 33.05 \\
                       & MF-NeRF-8 ($2^{20}$)   & 35.55 & 25.89 & 34.21 & 37.41 & 35.98 & 29.54 & 35.4  & 30.49 & 33.06 \\
                       & MF-NeRF-8 ($2^{22}$)   & 35.74 & 25.95 & 34.03 & 37.46 & 36.27 & 29.58 & 35.86 & 30.71 & 33.20 \\ \midrule
\multirow{7}{*}{SSIM}  & TensoRF                & .9845 & .9363 & .9827 & .9828 & .9835 & .9523 & .9886 & .8938 & .9630 \\
                       & DVGOv1                 & .9766 & .9304 & .9785 & .9799 & .9756 & .9509 & .9828 & .8788 & .9567 \\
                       & DVGOv2                 & .9758 & .9292 & .9774 & .9801 & .9763 & .9503 & .9827 & .8771 & .9561 \\
                       & Instant-NGP ($2^{20}$) & .9800 & .9313 & .9813 & .9804 & .9798 & .9450 & .9888 & .8891 & .9595 \\
                       & Instant-NGP ($2^{22}$) & .9819 & .9326 & .9821 & .9805 & .9807 & .9453 & .9894 & .8891 & .9602 \\
                       & MF-NeRF-8 ($2^{20}$)   & .9814 & .9320 & .9826 & .9807 & .9800 & .9454 & .9878 & .8901 & .9600 \\
                       & MF-NeRF-8 ($2^{22}$)   & .9810 & .9338 & .9818 & .9813 & .9813 & .9465 & .9895 & .8894 & .9606 \\ \midrule
\multirow{7}{*}{LPIPS} & TensoRF                & .0219 & .0719 & .0220 & .0304 & .0178 & .0584 & .0146 & .1386 & .0470 \\
                       & DVGOv1                 & .0269 & .0769 & .0239 & .0340 & .0282 & .0578 & .0173 & .1619 & .0534 \\
                       & DVGOv2                 & .0276 & .0797 & .0254 & .0336 & .0269 & .0586 & .0177 & .1615 & .0539 \\
                       & Instant-NGP ($2^{20}$) & .0223 & .0781 & .0262 & .0360 & .0211 & .0677 & .0173 & .1318 & .0501 \\
                       & Instant-NGP ($2^{22}$) & .0212 & .0765 & .0255 & .0354 & .0198 & .0659 & .0162 & .1253 & .0482 \\
                       & MF-NeRF-8 ($2^{20}$)   & .0216 & .0772 & .0238 & .0346 & .0215 & .0687 & .0189 & .1333 & .0499 \\
                       & MF-NeRF-8 ($2^{22}$)   & .0210 & .0754 & .0255 & .0330 & .0193 & .0653 & .0158 & .1284 & .0480 \\ \midrule
\multirow{7}{*}{\begin{tabular}[c]{@{}c@{}}Training\\ Time\end{tabular}} &
  TensoRF &
  23.41 &
  24.15 &
  29.18 &
  29.40 &
  26.91 &
  40.70 &
  19.94 &
  36.48 &
  28.77 \\
                       & DVGOv1                 & 17.27 & 16.95 & 18.35 & 21.43 & 18.23 & 21.32 & 15.27 & 23.37 & 19.02 \\
                       & DVGOv2                 & 7.98  & 7.90  & 8.03  & 9.05  & 8.55  & 9.23  & 6.85  & 11.03 & 8.58  \\
                       & Instant-NGP ($2^{20}$) & 7.331 & 6.925 & 5.585 & 9.754 & 8.56  & 7.93  & 5.59  & 15.7  & 8.42  \\
                       & Instant-NGP ($2^{22}$) & 11.88 & 11.13 & 8.76  & 15.79 & 13.82 & 12.59 & 8.85  & 25.61 & 13.55 \\
                       & MF-NeRF-8 ($2^{20}$)   & 7.29  & 6.74  & 5.06  & 9.77  & 8.39  & 8.04  & 5.17  & 16.47 & 8.37  \\
                       & MF-NeRF-8 ($2^{22}$)   & 10.47 & 9.72  & 7.12  & 14.00 & 12.36 & 11.34 & 7.16  & 25.27 & 12.18 \\ \bottomrule
\end{tabular}
\caption{Scene by scene results on a synthetic dataset (Synthetic NeRF).}
\label{tab:1}
\end{table*}

\begin{table*}[t]
\centering
\begin{tabular}{@{}c|l|ccccccc|c@{}}
\toprule
Metrics &
  \multicolumn{1}{c|}{Methods} &
  Bicycle &
  Bonsai &
  Counter &
  Garden &
  Kitchen &
  Room &
  Stump &
  Avg. \\ \midrule
\multirow{5}{*}{PSNR}  & DVGOv2                & 22.09 & 28.27 & 26.13 & 24.38 & 26.49 & 28.97 & 23.57 & 25.70 \\
                       & Instant-NGP ($2^{20}$) & 23.21 & 30.80 & 26.88 & 24.95 & 30.27 & 29.94 & 23.39 & 27.06 \\
                       & Instant-NGP ($2^{22}$) & 23.39 & 31.29 & 27.16 & 25.44 & 30.52 & 30.08 & 23.62 & 27.36 \\
                       & MF-NeRF-8 ($2^{20}$)   & 23.64 & 30.87 & 26.95 & 24.7  & 30.18 & 29.96 & 23.48 & 27.11 \\
                       & MF-NeRF-8 ($2^{22}$)   & 23.50 & 31.46 & 27.4  & 25.29 & 30.66 & 30.25 & 23.12 & 27.38 \\ \midrule
\multirow{5}{*}{SSIM}  & DVGOv2                & .4798 & .8444 & .7861 & .6297 & .7373 & .8714 & .5816 & .7043 \\
                       & Instant-NGP ($2^{20}$) & .5263 & .9144 & .8036 & .6489 & .8831 & .8947 & .5474 & .7455 \\
                       & Instant-NGP ($2^{22}$) & .5543 & .9246 & .8250 & .6998 & .8983 & .9041 & .5833 & .7699 \\
                       & MF-NeRF-8 ($2^{20}$)   & .5308 & .9126 & .7979 & .6238 & .8755 & .8940 & .5451 & .7400 \\
                       & MF-NeRF-8 ($2^{22}$)   & .5508 & .9256 & .8278 & .6833 & .8993 & .9046 & .5573 & .7641 \\ \midrule
\multirow{5}{*}{LPIPS} & DVGOv2                & .5107 & .2817 & .3191 & .3765 & .3057 & .3054 & .4586 & .3654 \\
                       & Instant-NGP ($2^{20}$) & .5122 & .1971 & .3020 & .3403 & .1754 & .2578 & .4855 & .3243 \\
                       & Instant-NGP ($2^{22}$) & .4814 & .1705 & .2715 & .2984 & .1536 & .2312 & .4510 & .2939 \\
                       & MF-NeRF-8 ($2^{20}$)   & .5091 & .2058 & .3142 & .3621 & .1846 & .2675 & .4960 & .3342 \\
                       & MF-NeRF-8 ($2^{22}$)   & .4884 & .1721 & .2707 & .3145 & .1543 & .2374 & .4736 & .3016 \\ \midrule
\multirow{5}{*}{\begin{tabular}[c]{@{}c@{}}Training\\ Time\end{tabular}} &
  DVGOv2 &
  27.10 &
  23.68 &
  24.55 &
  24.18 &
  21.92 &
  23.42 &
  26.03 &
  24.41 \\
                       & Instant-NGP ($2^{20}$) & 11.68 & 11.35 & 13.23 & 12.86 & 10.22 & 8.50  & 19.26 & 12.44 \\
                       & Instant-NGP ($2^{22}$) & 22.62 & 22.40 & 22.27 & 24.70 & 18.63 & 16.13 & 32.91 & 22.81 \\
                       & MF-NeRF-8 ($2^{20}$)   & 10.02 & 12.13 & 12.48 & 13.62 & 9.85  & 8.74  & 15.61 & 11.78 \\
                       & MF-NeRF-8 ($2^{22}$)   & 19.16 & 20.49 & 20.43 & 21.90 & 16.82 & 15.48 & 27.87 & 20.31 \\ \bottomrule
\end{tabular}
\caption{Scene by scene results on a unbounded dataset (MipNeRF360).}
\label{tab:2}
\end{table*}

\clearpage
\bibliography{references}